\documentclass[10pt,twocolumn,letterpaper]{article}

\usepackage{3dv}
\usepackage{times}
\usepackage{epsfig}
\usepackage{graphicx}
\usepackage{amsmath}
\usepackage{amssymb}

\usepackage{booktabs}
\usepackage{siunitx}
\usepackage{cite}
\usepackage{ctable}
\usepackage{hhline}
\usepackage{booktabs} 
\usepackage{subcaption}
\usepackage{csquotes}
\usepackage{multirow}
\usepackage{pifont}
\usepackage[super]{nth}
\usepackage{wrapfig}
\usepackage[normalem]{ulem}
\usepackage{subcaption}
\usepackage{transparent}
\usepackage{algorithm}
\usepackage{listings}
\usepackage[pagebackref=true,breaklinks=true,colorlinks,bookmarks=false]{hyperref}

\usepackage{cleveref}

\newcommand{\normsq}[1]{\Big\lVert#1\Big\rVert^2_2}
\usepackage{etoolbox}
\makeatletter
\AfterEndEnvironment{algorithm}{\let\@algcomment\relax}
\AtEndEnvironment{algorithm}{\kern2pt\hrule\relax\vskip3pt\@algcomment}
\let\@algcomment\relax
\newcommand\algcomment[1]{\def\@algcomment{\footnotesize#1}}
\renewcommand\fs@ruled{\def\@fs@cfont{\bfseries}\let\@fs@capt\floatc@ruled
  \def\@fs@pre{\hrule height.8pt depth0pt \kern2pt}%
  \def\@fs@post{}%
  \def\@fs@mid{\kern2pt\hrule\kern2pt}%
  \let\@fs@iftopcapt\iftrue}
\makeatother

\newcommand{\cmark}{\ding{51}}%
\newcommand{\xmark}{\ding{55}}%

\newcolumntype{L}[1]{>{\raggedright\let\newline\\\arraybackslash\hspace{0pt}}m{#1}}
\newcolumntype{C}[1]{>{\centering\let\newline\\\arraybackslash\hspace{0pt}}m{#1}}
\newcolumntype{R}[1]{>{\raggedleft\let\newline\\\arraybackslash\hspace{0pt}}m{#1}}

\makeatletter
\newcommand*{\rom}[1]{\expandafter\romannumeral #1}
\def\blfootnote{\xdef\@thefnmark{}\@footnotetext}
\makeatother

\threedvfinalcopy 

\def\threedvPaperID{124} 

\ifthreedvfinal\fi

\begin{document}

\title{MonoNHR: Monocular Neural Human Renderer}

\thispagestyle{empty}

\author{
Hongsuk Choi$^{*1,3}$
\and
Gyeongsik Moon$^{*2}$
\and
Matthieu Armando$^3$
\and
Vincent Leroy$^3$
\and
Kyoung Mu Lee$^1$
\and
Gr\'egory Rogez$^3$
\and
\\
$^1$ Dept. of ECE \& ASRI, Seoul National University, Korea
\\
$^2$ Meta Reality Labs Research
\hspace{1.1cm}
$^3$ NAVER LABS Europe
\\
\small \texttt {redstonepo@gmail.com, mks0601@fb.com, kyoungmu@snu.ac.kr}\\ \small \texttt {\{matthieu.armando,vincent.leroy,gregory.rogez\}@naverlabs.com}
}

\maketitle
\begin{abstract}
Existing neural human rendering methods struggle with a single image input due to the lack of information in invisible areas and the depth ambiguity of pixels in visible areas.
In this regard, we propose Monocular Neural Human Renderer (MonoNHR), a novel approach that renders robust free-viewpoint images of an arbitrary human given only a single image.
MonoNHR is the first method that (i) renders human subjects never seen during training in a monocular setup, and (ii) is trained in a weakly-supervised manner without geometry supervision.
First, we propose to disentangle 3D geometry and texture features and to condition the texture inference on the 3D geometry features.
Second, we introduce a~\textit{Mesh Inpainter} module that inpaints the occluded parts exploiting human structural priors such as symmetry.
Experiments on ZJU-MoCap, AIST and HUMBI datasets show that our approach significantly outperforms the recent methods adapted to the monocular case.
\blfootnote{* equal contribution}
\end{abstract}

\begin{figure}[!t]
\begin{center}
\includegraphics[width=0.22\linewidth]{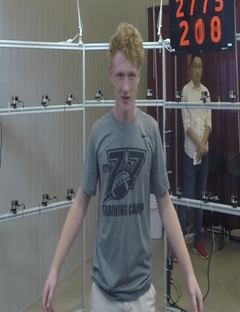}
\includegraphics[width=0.22\linewidth]{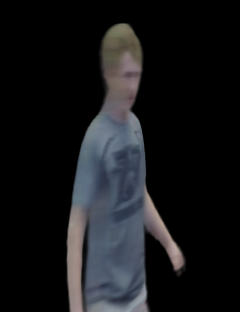}
\includegraphics[width=0.22\linewidth]{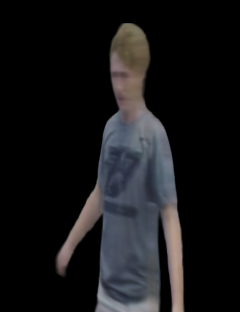}
\includegraphics[width=0.22\linewidth]{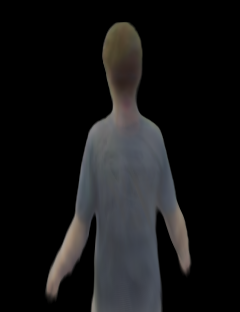} \\
\includegraphics[width=0.22\linewidth]{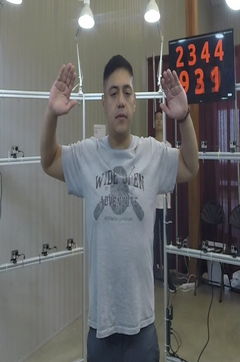} 
\includegraphics[width=0.22\linewidth]{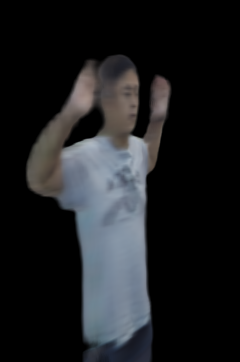}
\includegraphics[width=0.22\linewidth]{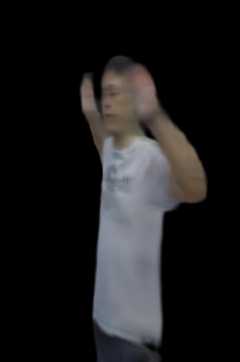}
\includegraphics[width=0.22\linewidth]{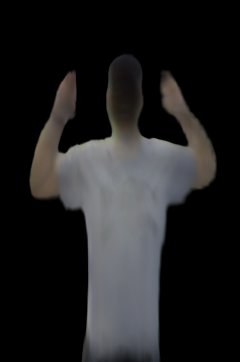} \\
\includegraphics[width=0.22\linewidth]{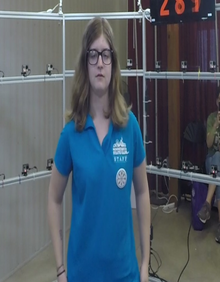}
\includegraphics[width=0.22\linewidth]{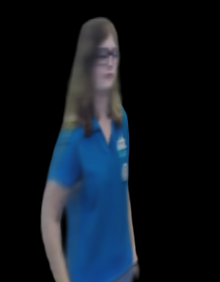}
\includegraphics[width=0.22\linewidth]{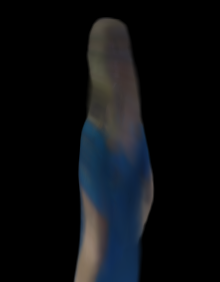}
\includegraphics[width=0.22\linewidth]{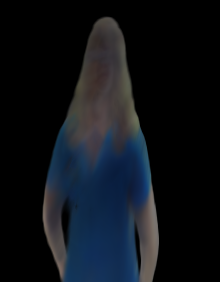} \\
\end{center}
\vspace*{-5mm}
\caption{
Given a \textbf{single} input image of a human (leftmost), MonoNHR generates realistic renderings from novel viewpoints (right). Tested on unseen subjects from HUMBI~\cite{yu2020humbi}.
}
\vspace*{-5mm}
\label{fig:banner}
\end{figure}

\section{Introduction}~\label{sec:intro}

Novel view synthesis \textbf{from a single image} is a very challenging problem, but has many potential applications,~\eg in AR/VR or smartphone-based social networking services. Markerless capture from RGB data has been widely studied as a tool to generate realistic free-viewpoint renderings of humans, but it often requires synchronized and calibrated multi-camera systems. We take a step towards monocular capture of people's appearance and shape, and tackle novel view synthesis of a person observed from a single image, which extends concurrent works to a more general setting.

Neural human rendering methods, which aim to render people from arbitrary viewpoints, showed promising results for this task. These can generally be grouped into 2 main categories: those learning subject-specific Neural Radiance Fields (NeRF)~\cite{mildenhall2020nerf} to represent the appearance of a particular human~\cite{wu2020multi,liu2021neural,peng2021neural,saito2021scanimate,peng2021animatable, su2021anerf} and approaches that estimate neural surface fields~\cite{shao2021doublefield} using pixel-aligned image features~\cite{saito2019pifu,saito2020pifuhd,hong2021stereopifu,zheng2021pamir,shao2021doublefield}. The first ones require a large number of input images~\cite{wu2020multi,liu2021neural} or multi-view video frames capturing the complete surface of the target~\cite{wu2020multi,liu2021neural,peng2021neural,peng2021animatable}, while the others rely on a detailed geometric ground-truth during training, and thus, require expensive, therefore small-scale, 3D scans datasets, preventing generalization to unseen human poses and appearances. ARCH~\cite{huang2020arch} and ARCH++~\cite{he2021arch}, that follow a different approach by learning an occupancy function in some canonical body space, also suffer from this problem. We summarize these modalities in Table~\ref{table:intro_compare}.

Interestingly, very recent NeRF-based methods, such as pixelNeRF~\cite{yu2021pixelnerf}, PVA~\cite{raj2021pixel}, and NHP~\cite{kwon2021neural}, showed that it is feasible to render free-viewpoint images of humans from a sparse set of views, while allowing generalization to arbitrary subjects. However, these methods are not designed to synthesize occluded surfaces, as they only render surfaces visible in the input views.
Thus these methods struggle with monocular inputs, where more than half the surface of the observed person can be invisible to the camera, making the texture and geometry of the invisible parts largely ambiguous. Furthermore, a depth ambiguity remains inherent to monocular observations. 
In this paper, we claim that prior knowledge of the human appearance and shape, such as symmetry, color consistency between surfaces, and front-back coherence, should be better exploited for this task.  

Based on these observations, we propose Monocular Neural Human Renderer (MonoNHR), a novel NeRF-based architecture that robustly renders free-viewpoint images of an arbitrary human given a single image of that person. We address the issues inherent to a monocular observation in two ways.
First, we disentangle the features of 3D geometry and texture by extracting \emph{geometry-dedicated features}.
Different from neural surface field-based methods~\cite{saito2019pifu,saito2020pifuhd}, MonoNHR is trained only with multi-view images without ground-truth (GT) 3D scans.
Since we do not consider explicit 3D geometry, extracting geometry-dedicated features is non-trivial.
To do so, we design a geometry estimation branch, separated from the texture estimation fork, that is used solely for estimating the density of the radiance field. Second, we introduce ~\textit{Mesh Inpainter} that operates on the SMPL~\cite{loper2015smpl} mesh estimated from the input image. It is used only during training, to encourage the backbone network to implicitly learn human priors. Please note that contrary to SMPL texturing works~\cite{xu2021texformer}, our method does not rely on this 3D surface and is able to render shapes that largely differ from the SMPL model.

We study the efficacy of MonoNHR on ZJU-MoCap~\cite{peng2021neural}, AIST~\cite{li2021learn,tsuchida2019aist} and HUMBI~\cite{yu2020humbi} datasets.
Experiments show that our method significantly outperforms recent NeRF-based methods 
on monocular images.
To the best of our knowledge, MonoNHR is the first approach specifically designed for novel view synthesis of humans from a~\textit{monocular} image using neural radiance fields. Unlike previous works, it explicitly and effectively handles the many ambiguities inherent to monocular observations. Our contributions are summarized below:

\begin{itemize}
\item We present MonoNHR, a novel NeRF-based architecture that robustly renders free-viewpoint images of an arbitrary human from a monocular image. 
It pushes the boundaries of NeRF-based novel view synthesis research to a more general setting.
\item We design the network to handle the specific challenges of the task, such as invisible surface synthesis and depth ambiguity. We tackle the former via a mesh inpainting module. For the latter, we disentangle 3D geometry and texture features, and condition texture inference based on the geometry features.
\item The proposed system significantly outperforms previous methods on monocular images both quantitatively and qualitatively, and achieves state-of-the-art rendering quality on novel view synthesis benchmarks. 
\end{itemize}

\begin{table}[t]
\footnotesize
\centering
\setlength\tabcolsep{1.0pt}
\def\arraystretch{1.1}
\begin{tabular}{C{3.0cm}|C{2.0cm}C{2.0cm}C{1.0cm}}
\specialrule{.1em}{.05em}{.05em}
method & input & supervision & unseen identity \\ \hline
NB~\cite{peng2021neural}, Ani-NeRF~\cite{peng2021animatable} & subject code & multi-view videos & \textcolor[RGB]{252,65,37}{\xmark} \\
PIFu~\cite{saito2019pifu,saito2020pifuhd} ARCH/ARCH++~\cite{huang2020arch,he2021arch} & \textbf{monocular image} & 3D scans & \textcolor[RGB]{21,165,63}{\cmark} \\
NHP~\cite{kwon2021neural}& multi-view videos & \textbf{multi-view images} & \textcolor[RGB]{21,165,63}{\cmark} \\ 
\textbf{MonoNHR (Ours)} & \textbf{monocular image} & \textbf{multi-view images} & \textcolor[RGB]{21,165,63}{\cmark} \\ \specialrule{.1em}{.05em}{.05em}
\end{tabular}
\vspace*{-3mm}
\caption{Comparison of recent neural human rendering methods.
MonoNHR is the first work that 1) takes a monocular image as an input, 2) is supervised with multi-view images without 3D scans, and 3) is generalizable to unseen subjects (identities).
}
\label{table:intro_compare}
\end{table}
\section{Related work}~\label{sec:related_work}
\vspace{-3mm}

\noindent\textbf{Markerless Performance Capture from RGB data} often required large acquisition platforms with tens to hundreds of cameras~\cite{theobalt10performance,leroy21volume,collet15streamable}. The problem has also been approached through the use of sparse setups~\cite{huang18deepvolumetric,wu13onset}.  
We refer the reader to~\cite{xia17survey} for a broader overview. Nevertheless, all the multi-view approaches share the same limitations: synchronizing and calibrating multi-camera systems
is cumbersome, requires storing and processing large amounts of data, and not always feasible in practice.
A few recent approaches tackled the monocular case~\cite{habermann19livecap,xu20eventcap,xu18monoperfcap}, but they all require pre-scanned templates of the subjects. 

Furthermore, most of the performance capture methods~\cite{guo19relightables,collet15streamable,armando19adaptive,franco09efficient,starck07surface,ma21pixel} solve the problem in two distinct steps: 1) reconstructing the mesh of the observed subject, and 2) coloring it using available observations, possibly considering lighting information. The main drawback of such a strategy is that the appearance is conditioned on, but also limited by, the geometry, which is inherently noisy and inaccurate if not incomplete. 
In this work, we wish to switch paradigms and directly model view-dependent appearance. In fact, the idea was already introduced in~\cite{debevec00reflectance}. We would like to follow their work, by investigating the potential use of NeRF~\cite{mildenhall2020nerf} to represent view-dependent appearances without relying on explicit geometry formulation using only a single input view.

\noindent\textbf{NeRF-based  human rendering} methods implicitly encode a dense scene radiance field in the form of a density and a color for a given 3D query point and viewing direction via a neural network.
One of their advantages is that they do not require 3D supervision, and instead rely on 2D supervision from multi-view images.
The main drawback of the original formulation, however, is that a NeRF model has to be optimized for each scene, since it is, in fact, an optimization scheme, and it is not a learning approach \emph{per se}.
One stream of NeRF-based approaches for human rendering is building a subject-specific representation.
NHR~\cite{wu2020multi} takes a sequence of point clouds as an input and conditions the rendered novel images using 80 input points of view.
NB~\cite{peng2021neural} utilizes a 3D human mesh model (\ie SMPL~\cite{loper2015smpl}) and subject-specific latent codes, to construct the 3D latent code volume, which is used for density and color regression of any 3D point bounded by a given 3D human mesh. Other works
~\cite{chen2021animatable,peng2021animatable,liu2021neural} deform observation-space 3D points to the canonical 3D space using inverse bone transformations, and learn the neural radiance fields.
The canonical 3D space represents the pose normalized space around the template human mesh.
NARF~\cite{noguchi2021neural} learns neural radiance fields per human part, and trains an autoencoder to encode human appearances using a synthetic human dataset.


Recently, pixelNeRF~\cite{yu2021pixelnerf}, IBRNet~\cite{wang2021ibrnet}, and SRF~\cite{chibane2021stereo} proposed to combine an image-based feature encoding and NeRF.
Instead of memorizing the scene radiance in a 3D space, their networks estimate it based on pixel-aligned image features. 
We chose to adopt this strategy, as it allows a single network to be trained across multiple scenes to learn a scene prior, enabling it to generalize to unseen scenes in a feed-forward manner from a sparse set of views. 
PVA~\cite{raj2021pixel}, Wang~et al.~\cite{wang2021learning}, and NHP~\cite{kwon2021neural} also apply this approach to human rendering.
Especially, NHP targets human body rendering, which is also our interest, using sparse multi-view videos.
Given a GT SMPL mesh, it exploits the pixel-aligned image features to construct the 3D latent volume.

Among the above generalizable NeRF-extensions~\cite{raj2021pixel,wang2021learning,kwon2021neural}, no work has addressed the challenges inherent to single monocular images, such as occlusions and depth ambiguity.
Instead, they attempted to resolve the issues by increasing the input information, such as using more views and temporal information,
which in practice can be a limiting or even prohibitive factor for real-world applications.  
In this paper, we demonstrate the robustness of MonoNHR on single images by comparing it with pixelNeRF~\cite{yu2021pixelnerf} and NHP~\cite{kwon2021neural}, which are generalizable NeRF-extensions and are applicable to monocular images.


\noindent\textbf{Neural surface fields-based human rendering} methods are closely related to NeRF representations, since they also aim at learning an implicit function, in this case, an indicator of the interior and exterior of the observed shape. 
They allow for accurate and detailed 3D geometry reconstructions, but they require strong 3D supervision, such as 3D scans.
3D scans are highly costly to obtain at scale, and consequently, methods are trained on a small scan dataset and tend to exhibit poor generalization capabilities to unseen human poses and appearances. 
PIFu~\cite{saito2019pifu} and its extensions~\cite{saito2020pifuhd,hong2021stereopifu} propose to estimate the 3D surface of a human using an implicit function based on \textit{pixel-aligned} image features.
The \textit{pixel-aligned} image features are obtained by projecting 3D points onto the image plane. 
Similar to more classical approaches, they first reconstruct 3D surfaces and then condition texture inference on surface reconstruction features similar to ours. 
However, as DoubleField~\cite{shao2021doublefield} pointed out, the learning space of texture is highly limited around the surface and discontinuous, which hinders the optimization.
Zins~et al.~\cite{zins2021learning} improves PIFu in a multi-view setting by introducing an attention-based view fusion layer and a context encoding module using 3D convolutions.
POSEFusion~\cite{li2021posefusion} takes monocular RGBD video frames as an input and learns to fuse multiple surface estimations in different time steps.
DoubleField jointly learns neural radiance and surface fields, and uses raw RGB pixel values to render high-resolution images.

Compared to the above neural surface fields-based human rendering methods, MonoNHR has two clear differences.
First, it does not require 3D scans for training 
and follows NeRF-based human rendering pipeline, \ie using a weak supervision signal from multi-view images. 
Although NeRF-based human rendering methods, including ours, use SMPL fits, the SMPL fits are much easier to obtain than 3D scans using existing powerful 3D human pose and shape estimation methods~\cite{kolotouros2019learning,moon2022hand4whole}.
On the other hand, special and expensive equipments, \eg over 100 multi-view synchronized cameras, are necessary to generate accurate 3D scans, making them difficult to obtain at a large scale.
Please note that 3D scans obtained from a small number of cameras, \eg COLMAP~\cite{schonberger2016structure,schonberger2016pixelwise}, are not accurate enough to provide supervision targets for PIFu and its variants, as discussed in ~\cite{peng2021neural,leroy2018shape}.
In consequence, PIFu and its variants are trained on small scale datasets and tend not to generalize well to unseen data, especially non-upright standing poses, as discussed in~\cite{kwon2021neural,peng2021neural}.
Second, the absence of 3D scans prevents explicit 3D geometry supervision, making disentangling geometry and texture non-trivial.
We extract geometry-dedicated features for disentanglement and use them for density estimation without RGB estimation.




\begin{figure*}[!t]
\begin{center}
\includegraphics[width=0.9\linewidth]{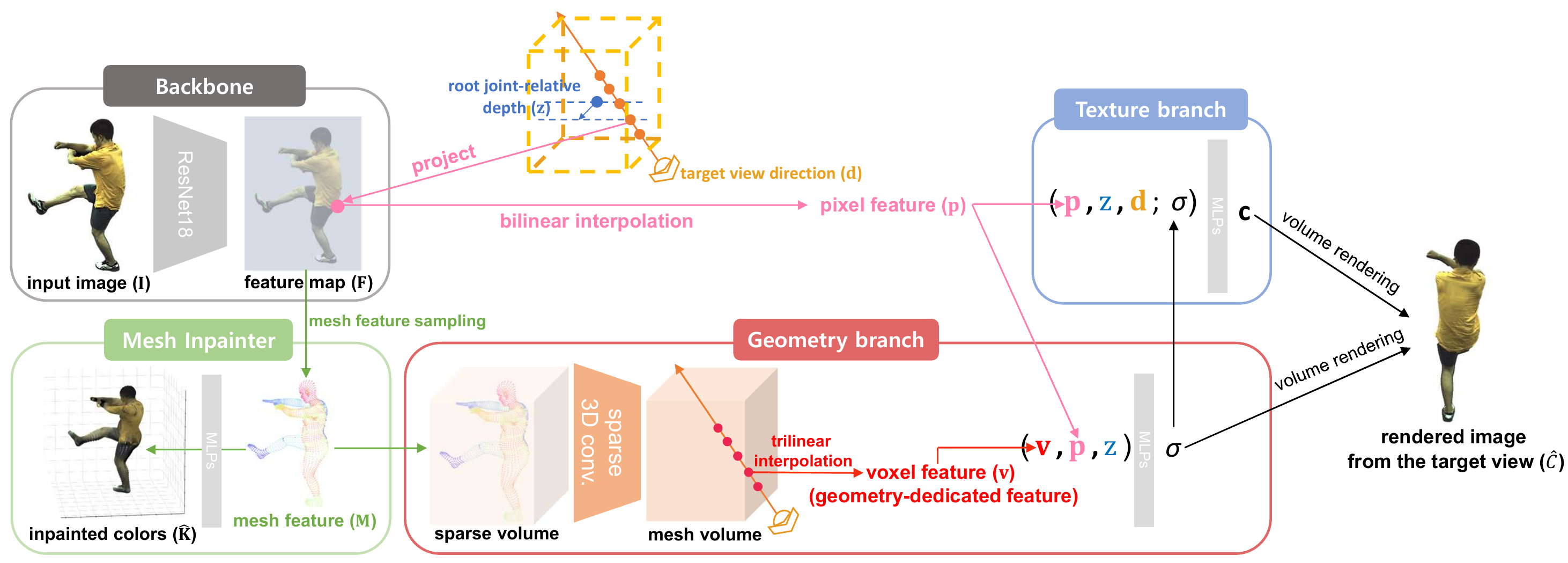}
\end{center}
\vspace{-5mm}
\caption
{MonoNHR overall pipeline.
The backbone produces a feature map $\mathbf{F}$ from a single input image $\mathbf{I}$.
Then, \textcolor[RGB]{252,99,159}{pixel features $\mathbf{p}$} and \textcolor[RGB]{98,163,59}{mesh feature $\mathbf{M}$} are extracted from  $\mathbf{F}$.
Mesh Inpainter is used to predict colors of all mesh vertices $\hat{\mathbf{K}}$.
The geometry branch takes the mesh features, processes them using sparse 3D CNN, and then extracts \textcolor[RGB]{252,22,39}{voxel features $\mathbf{v}$}.
From a combination of voxel features, pixel features, and  \textcolor[RGB]{54,93,183}{root joint-relative depths $\mathrm{z}$}, the geometry branch outputs a density value.
Finally, the texture branch predicts a RGB value from pixel features, root joint-relative depth, \textcolor[RGB]{227,142,24}{target view's ray direction $\mathbf{d}$}, and the predicted density value.
We render the target view $\hat{C}$ via volumetric rendering.
}
\label{fig:pipeline}
\end{figure*}

\section{MonoNHR}~\label{sec:mono_nhr}
\vspace{-1em}

The overall pipeline of MonoNHR is detailed in Figure~\ref{fig:pipeline}.
It is trained in an end-to-end manner and consists of an image feature backbone, Mesh Inpainter, a geometry branch, and a texture branch.

\subsection{Backbone}
First, a backbone extracts image features that will be used to condition the neural radiance field. Similarly to~\cite{yu2021pixelnerf}, this allows MonoNHR to learn priors relevant to human rendering, which ultimately enables our method to work on subjects that were not seen during training.
We use a ResNet18~\cite{he2016deep} as backbone.
It takes a masked human image $\mathbf{I} \in \mathbb{R}^{3 \times 256 \times 256}$ as input and produces a feature map $\mathbf{F} \in \mathbb{R}^{c_p \times 128 \times 128}$, with
$c_p=64$ the feature's channel dimension. An off-the-shelf semantic part segmentation method, PGN~\cite{gong2018instance}, is used to segment the subjects in the images.
The feature map $\mathbf{F}$ is used in two ways:

\noindent\textbf{Mesh feature.} Following NHP~\cite{kwon2021neural}, we sample per-vertex image features from $\mathbf{F}$, given SMPL~\cite{loper2015smpl} mesh vertices.
We project these 3D vertices onto the image plane and use bilinear interpolation to sample the corresponding image features. 
Then, we concatenate the per-vertex image features and the vertices' root joint (~\textit{i.e.}, pelvis)-relative depths, to distinguish the features of both occluded and occluding parts.
The concatenated feature is called the \textcolor[RGB]{98,163,59}{mesh feature} and denoted as \textcolor[RGB]{98,163,59}{$\mathbf{M} \in \mathbb{R}^{M \times (c_p+1)}$}, where $M$ denotes the number of SMPL mesh vertices. We feed these to the geometry branch.

\noindent\textbf{Pixel feature.} We sample 3D query points from a ray of a target view (rendering view), following the quadrature rule of the volume rendering discussed by Max~\cite{max1995optical}. The rays are bounded by the given mesh's 3D bounding box, following NB~\cite{peng2021neural} and NHP~\cite{kwon2021neural}. 
The 3D query point $\mathbf{x}\in \mathbb{R}^3$ is equipped with appearance information from its projection on the image plane. We coin this information the \textcolor[RGB]{252,99,159}{pixel feature $\mathbf{p} \in \mathbb{R}^{c_p}$}.
The pixel feature $\mathbf{p}$ is fed to both geometry and texture branches with $\mathbf{x}$'s \textcolor[RGB]{54,93,183}{root joint-relative depth $\mathrm{z}$}.
The root joint-relative depth is defined in the input view's camera-centered coordinate system.

\subsection{Mesh Inpainter}~\label{sec:mesh_inpainter}
This module reconstructs the colors of mesh vertices from a monocular image. 
It is a simple MLP network that consists of two layers. 
It takes the mesh feature $\mathbf{M}$ as input and regresses the vertices' colors $\hat{\mathbf{K}} \in \mathbb{R}^{M \times 3}$.
$\hat{\mathbf{K}}$ is only used during training.
The vertices' GT colors $\mathbf{K}$ are obtained by 1.
projecting the vertices to multi-view images considering visibility, 2. sampling the colors from each image with bilinear interpolation and 3. averaging the visible vertices' sampled colors.
The visibility of vertices are obtained by rasterizing mesh vertices.

Mesh Inpainter encourages the backbone network to implicitly learn human priors, such as symmetry of human parts and color similarity between surfaces belonging to the same part. 
While these could be learned by the rendering loss at the end of the network, we aim to facilitate it by giving a more direct signal to the backbone using mesh inpainting.
Mesh Inpainter's motivation is further explained in Section E of the supplementary material.

\subsection{Geometry branch}
The geometry branch regresses a density value $\sigma$ of a 3D query point $\mathbf{x}$. To this aim, we leverage sparse 3D convolutions similarly to NB~\cite{peng2021neural} and NHP~\cite{kwon2021neural}.

\noindent\textbf{Sparse 3D convolution.} 
We work with a sparse 3D volume, in which lie the sparse features of $\mathbf{M}$.
This sparse 3D volume is fed to a sparse 3D convolutional neural network (CNN)~\cite{graham2018spconv} that 
extracts multi-scale feature volumes, denoted mesh volumes.
From each mesh volume, we sample the feature corresponding to $\mathbf{x}$ based on its 3D camera coordinates using trilinear interpolation to get the \textcolor[RGB]{252,22,39}{voxel feature $\mathbf{v} \in \mathbb{R}^{c_v}$}, where
$c_v=192$ is the feature's channel dimension.
The voxel feature $\mathbf{v}$ is used only for the geometry estimation; therefore, it is a \emph{geometry-dedicated feature}.

\noindent\textbf{MLP.} 
Then, the density of $\mathbf{x}$ is predicted as a function of the voxel feature $\mathbf{v}$, the pixel feature $\mathbf{p}$, and the root joint-relative depth $\mathrm{z}$, as follows:
\begin{equation}
    \sigma(\mathbf{x}) = M_{\sigma}(\mathbf{v}, \mathbf{p}, \mathrm{z}),
\end{equation}
where $M_{\sigma}$ is another Multi-Layer Perceptron (MLP) network with four layers.

\subsection{Texture branch}
The texture branch regresses a RGB value $\mathbf{c}$ for a query point $\mathbf{x}$.
The RGB color of $\mathbf{x}$ is estimated as a function of the pixel feature $\mathbf{p}$, the root joint-relative depth $\mathrm{z}$, and the \textcolor[RGB]{227,142,24}{target view's ray direction $\mathbf{d} \in \mathbb{R}^{3}$}, conditioned on the predicted density value $\sigma$, as follows:
\begin{equation}
    \mathbf{c}(\mathbf{x}) = M_{\mathbf{c}}(\mathbf{p}, \mathrm{z}, \mathbf{R}\mathbf{d}; \sigma),
\end{equation}
where $M_{\mathbf{c}}$ represents an MLP network with five layers, and $\mathbf{R}$ denotes the world-to-camera rotation matrix.
$\mathbf{R}$ transforms the view direction to the input image's camera coordinate system. 
$\sigma$ provides information about the occupancy of a given 3D query point.
Such occupancy information is highly useful as RGB values should exist on the occupied points.
Please note that although we provide $\sigma$ to the texture branch, the voxel feature $\mathbf{v}$ is still a geometry-dedicated feature as its role is to predict accurate density $\sigma$.

\subsection{Volume rendering}
Given a target view, we use a classical differentiable ray-marching algorithm~\cite{kajiya1984ray} to render the target image following NeRF~\cite{mildenhall2020nerf}.
Concretely, the final color of a pixel is computed as the integral of RGB values along the ray shot from the camera center $\mathbf{o} \in \mathbb{R}^{3}$, weighted by predicted volume densities.
The integral is approximated via stratified sampling~\cite{mildenhall2020nerf}, and we use the quadrature rule~\cite{max1995optical} to limit memory usage and predict continuous radiance fields in practice.
For each pixel, we sample along the ray $N$ query points$\{\mathbf{x}\}_{i=1}^{N}$, where $\mathbf{x}_i = (\mathbf{o} + z_i\cdot\mathbf{d})$ and $z_i \in [z_\text{near},z_\text{far}]$.
$z_\text{near}$ and $z_\text{far}$ are the absolute depths of the two intersections between the ray and the given SMPL mesh's 3D bounding box. 
Then, the pixel color of the ray is computed as below:
\begin{gather}
    \hat{C}(\mathbf{r}) = \sum_{i=1}^{N} T_i (1 - \exp(-\sigma(\mathbf{x}_i) \delta_i)) \mathbf{c}(\mathbf{x}_i) \\
    \text{and} \quad T_i = \exp(-\sum_{j=1}^{i-1} \sigma(\mathbf{x}_j) \delta_j),
\end{gather}
where $\delta_i = || \mathbf{x}_{i + 1} - \mathbf{x}_{i} ||_2$ is the distance between adjacent sampled points, and $\mathbf{r}$ is the camera ray. 
In practice, we set $N$ to 64 following~\cite{peng2021neural,kwon2021neural}.

\subsection{Loss functions}
We supervise MonoNHR with two loss functions.

\noindent\textbf{Inpainting loss.} The inpainting loss compares the predicted and GT colors of vertices.
It is defined as below:
\begin{equation}
    L_{\text{inpaint}} = \sum\limits \normsq{\hat{\mathbf{K}} - \mathbf{K}},
\end{equation}
where $\hat{\mathbf{K}}$ and $\mathbf{K}$ indicate the predicted and GT colors of vertices, respectively.
Section~\ref{sec:mesh_inpainter} describes how we obtain the GT color of vertices $\mathbf{K}$.

\noindent\textbf{Rendering loss.} The rendering loss compares the predicted pixel color from the volume rendering and the GT pixel color.
We randomly sample four views to calculate the loss.
It is defined as below: 
\begin{equation}
    L_{\text{render}} = \sum\limits_{\mathbf{r} \in \mathcal{R}} \normsq{\hat{C}(\mathbf{r}) - C(\mathbf{r})},
\end{equation}
where $\mathcal{R}$ is a set of 3D query points on camera rays passing through image pixels, and $C(\mathbf{r})$ means the GT pixel color.

MonoNHR is trained in an end-to-end manner, and the total loss is defined as:
\begin{equation}
L_{\text{total}} = L_{\text{render}} + \lambda L_{\text{inpaint}}, 
\end{equation}
where $\lambda$ is set to $10^{-3}$.

\section{Experiments}

\subsection{Datasets}
\noindent\textbf{Datasets.}
We train and test MonoNHR on ZJU-MoCap~\cite{peng2021neural}, AIST~\cite{li2021learn,tsuchida2019aist}, CAPE~\cite{ma2020learning}, and HUMBI~\cite{yu2018humbi}.
For ZJU-MoCap, we split 10 videos of 10 subjects (1 video per 1 subject) to 7 training videos and 3 testing videos.
For AIST, we follow the official training and testing splits.
The training set has 834 videos of 20 dancing subjects, and the testing set has 374 videos of 10 dancing subjects.
Please note that the testing splits of all datasets contain poses and identities never seen during training.
CAPE is used to compare 3D geometry with non-NeRF methods as the above two datasets do not provide GT 3D geometry.
We use the above datasets for numerical studies, for more meaningful comparisons with prior work, but we also train on HUMBI for qualitative evaluation, for its larger size and diversity (see section~\ref{subsec:qualitative_results}). More details about data settings can be found in the supplementary material.


\noindent\textbf{Evaluation metrics.}
We report peak signal-to-noise ratio (PSNR) and structural similarity index (SSIM) to evaluate the quality of novel view synthesis, following~\cite{peng2021neural,yu2021pixelnerf,kwon2021neural}.
PSNR has been widely used to assess the quality of digital images. SSIM is designed based on luminance, contrast, and structure to better suit the workings of the human visual system~\cite{setiadi2021psnr}. 
For the 3D geometry evaluation, we report Chamfer distance (CD) in millimeters.


\begin{figure}[t]
\begin{center}
\includegraphics[width=\linewidth]{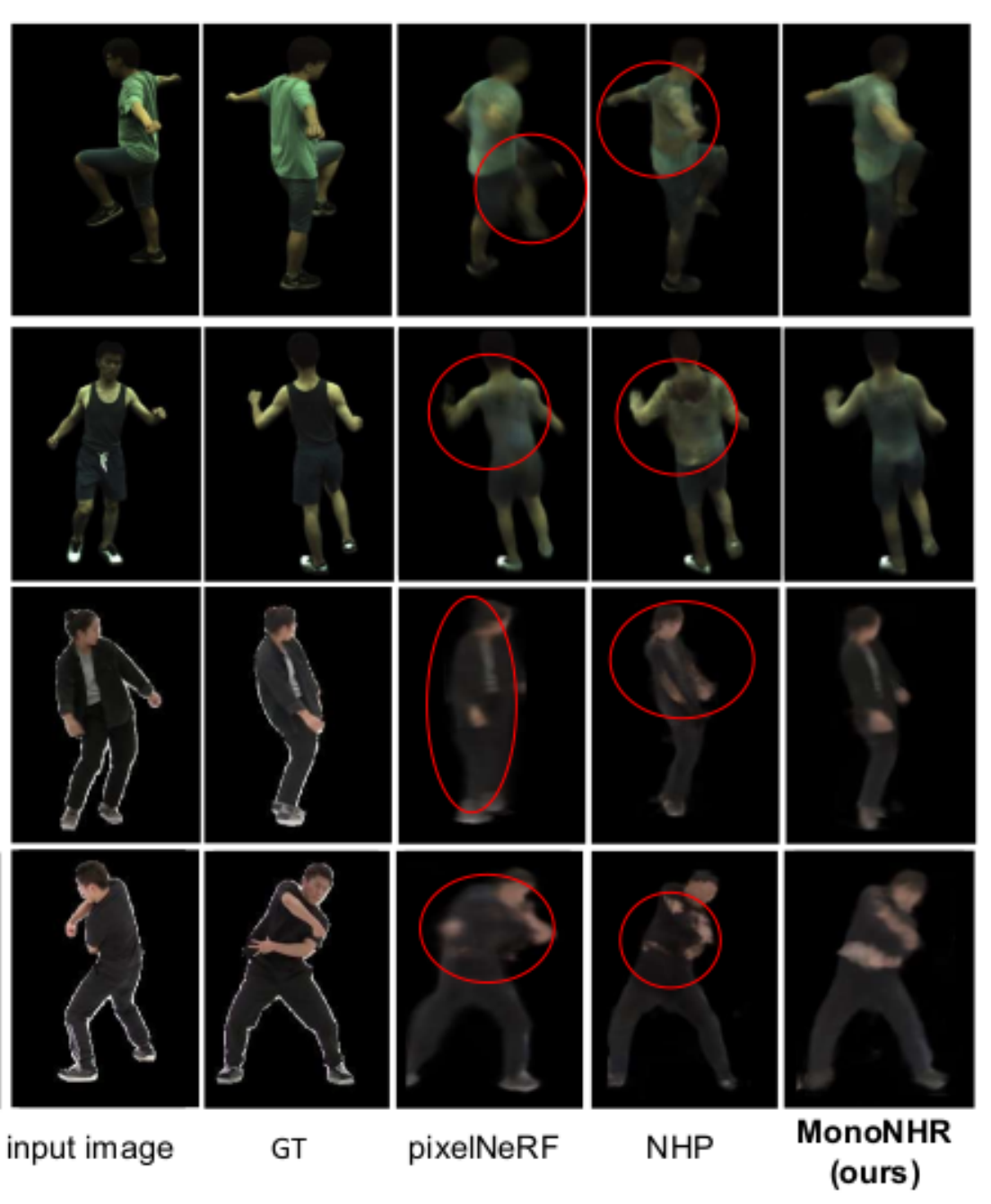} 
\end{center}
\vspace*{-5mm}
\caption{
Comparison of pixelNeRF~\cite{yu2021pixelnerf}, NHP~\cite{kwon2021neural}, and MonoNHR.
All results are obtained from monocular images with unseen poses and identities.
Top two rows: results on ZJU-MoCap, bottom two rows: results on AIST.
}
\vspace*{-3mm}
\label{fig:sota}
\end{figure}

\begin{figure*}[t]
\begin{center}
\includegraphics[width=\linewidth]{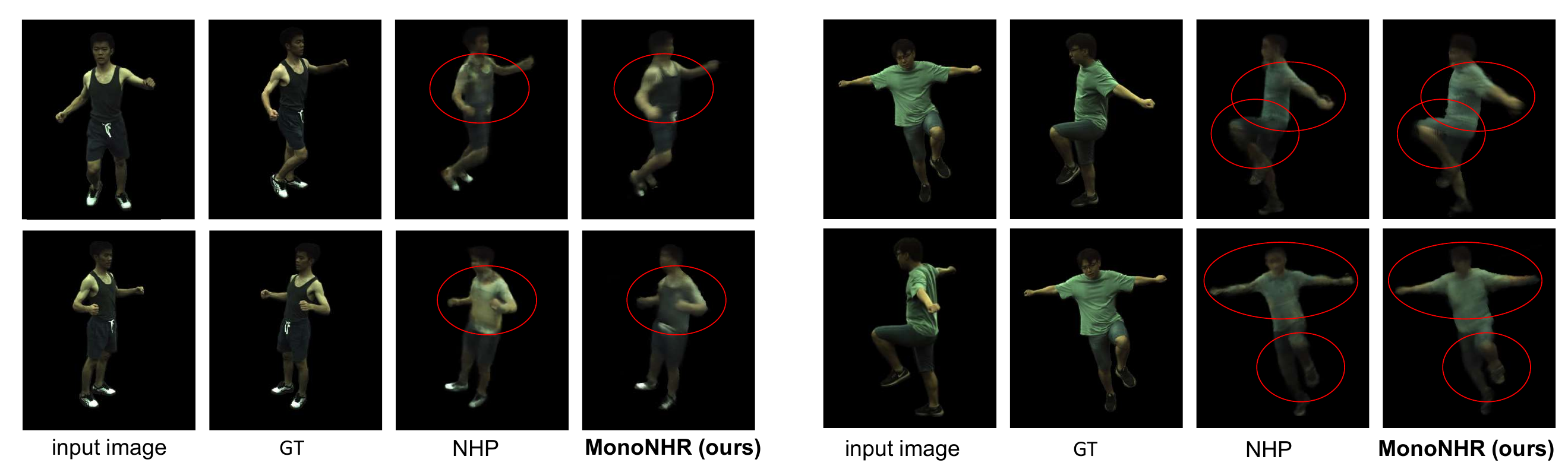}
\end{center}
\vspace*{-5mm}
\caption{
Comparison between NHP~\cite{kwon2021neural} and MonoNHR on ZJU-MoCap using estimated meshes of SPIN~\cite{kolotouros2019learning}.
}
\vspace*{-3mm}
\label{fig:estimated_sota}
\end{figure*}

\begin{figure*}[t]
\begin{center}
\includegraphics[width=\linewidth]{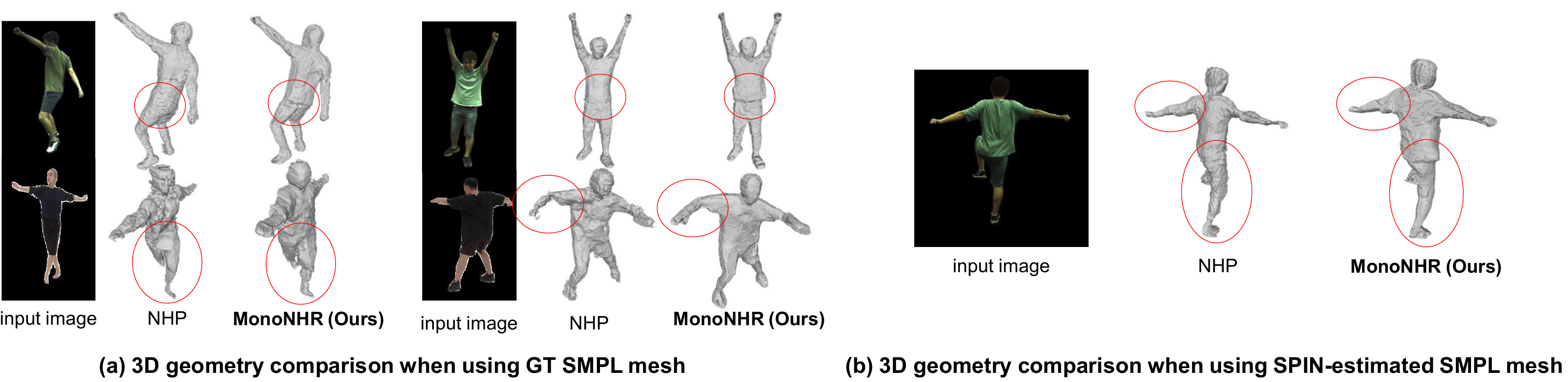}
\end{center}
\vspace*{-5mm}
\caption{
3D geometry visualization of NHP~\cite{kwon2021neural} and MonoNHR.
}
\vspace*{-2mm}
\label{fig:geo_comparison}
\end{figure*}

\subsection{Comparison with state-of-the-art methods}

\noindent\textbf{Overall comparison.}
The Figure~\ref{fig:sota} and Table~\ref{table:sota} show that MonoNHR produces the best results on both ZJU-MoCap~\cite{peng2021neural} and AIST~\cite{li2021learn,tsuchida2019aist}.

While pixelNeRF~\cite{yu2021pixelnerf} and NHP~\cite{kwon2021neural} tend to succeed in recovering only geometry or texture, MonoNHR produces robust novel view images in both aspects.
Concretely, pixelNeRF has a weakness in geometry reconstruction, and NHP struggles on recovering textures of invisible surfaces.
For example, on AIST, pixelNeRF has higher PSNR but lower SSIM than NHP.
PixelNeRF shows bent and missing human limbs 
and produces particularly unrecognizable structures of humans, when the target view is significantly different from the input view, and the human pose is complex.
NHP synthesizes better novel views than pixelNeRF in terms of geometry, but tends to exhibit noisy textures on invisible surfaces.
We think the noisy texture is due to the entanglement of geometry and texture features, which hinders a network from simultaneously optimizing them and eventually leads it to a local optimum that only optimizes either the geometry or the texture.
On the other hand, MonoNHR provides robust qualitative results of geometry and texture from diverse novel views, including the opposite view to the input image.
The superiority of MonoNHR not only comes from the disentanglement of geometry and texture features, but also from Mesh Inpainter that facilitates the system to learn human priors.
For instance, the first two rows of Figure~\ref{fig:sota} show that MonoNHR recovers robust geometry and texture of the subjects' occluded arms, which proves that it successfully learns human priors, such as symmetry of human parts, to compensate for the lack of information on invisible areas. 
We refer the reader to the supplementary material for additional comparisons.

\begin{table}[t]
\small
\centering
\setlength\tabcolsep{1.0pt}
\def\arraystretch{1.1}
\begin{tabular}{C{3.5cm}|C{1.1cm}C{1.1cm}|C{1.1cm}C{1.1cm}}
\specialrule{.1em}{.05em}{.05em}
\multirow{2}{*}{method} & \multicolumn{2}{c|}{ ZJU-MoCap} & \multicolumn{2}{c}{AIST} \\ 
&  PSNR $\uparrow$ &  SSIM $\uparrow$ & PSNR $\uparrow$ &  SSIM $\uparrow$ \\ \hline
pixelNeRF~\cite{yu2021pixelnerf} & 22.13 & 0.8604 & 16.79 & 0.6308\\
NHP~\cite{kwon2021neural} & 24.01 & 0.8953 & 16.58 & 0.6934\\ 
\textbf{MonoNHR (Ours)} & \textbf{25.36} & \textbf{0.9093} & \textbf{17.61} & \textbf{0.7186} \\ \specialrule{.1em}{.05em}{.05em}
\end{tabular}
\vspace*{-3mm}
\caption{
{Comparison of state-of-the-art NeRF-based neural human rendering methods on ZJU-MoCap and AIST.} 
}
\label{table:sota}
\end{table}

\begin{table}[t]
\small
\centering
\setlength\tabcolsep{1.0pt}
\def\arraystretch{1.1}
\begin{tabular}{C{3.0cm}|C{1.1cm}|C{1.1cm}}
\specialrule{.1em}{.05em}{.05em}
method & PSNR $\uparrow$ & SSIM $\uparrow$ \\ \hline
NHP~\cite{kwon2021neural} & 20.89 & 0.8364\\ 
\textbf{MonoNHR (Ours)} &  \textbf{22.25} & \textbf{0.8673}\\
\specialrule{.1em}{.05em}{.05em}
\end{tabular}
\vspace*{-3mm}
\caption{
{Comparison with NHP~\cite{kwon2021neural}, using estimated 3D human meshes of SPIN~\cite{kolotouros2019learning} on ZJU-MoCap.} 
}
\label{table:sota_mesh_estimate}
\end{table}

\noindent\textbf{Comparison using estimated 3D meshes.}
We also compare MonoNHR with NHP~\cite{kwon2021neural} on ZJU-MoCap~\cite{peng2021neural} using estimated body meshes of SPIN~\cite{kolotouros2019learning} in Table~\ref{table:sota_mesh_estimate}.
SPIN is a monocular method that regresses SMPL~\cite{loper2015smpl} parameters to produce a body mesh.
MonoNHR highly outperforms NHP, which also uses a body mesh at test time, considering that PSNR is in a log scale. 
Figure~\ref{fig:estimated_sota} further validates MonoNHR's robustness on  monocular images.
MonoNHR preserves the subject's 3D shape (\textit{e.g.}, loose clothes around the torso of the right-hand subject) while NHP relatively shrinks the subject's overall shape (\textit{e.g.}, the arms of the right-hand subject).
Also, the results of MonoNHR show more realistic textures in novel views than NHP (\textit{e.g.}, the shirt of the left-hand subject).


\noindent\textbf{3D geometry comparison.} 
Figure~\ref{fig:geo_comparison} compares the 3D geometry reconstruction of MonoNHR with NHP~\cite{kwon2021neural} on ZJU-MoCap.
MonoNHR clearly distinguishes the geometry of the top subjects' shirts and pants, while NHP shows vague boundaries between them. 
Also, on the SPIN example on the right (b), the result from MonoNHR reflects the widened deformation of the subject's shirt in the input image, but NHP recovers the shirt geometry in a tightened shape.
The results prove that MonoNHR learns better geometry features from a monocular image.

\begin{figure}[t]
\begin{center}
\includegraphics[width=\linewidth]{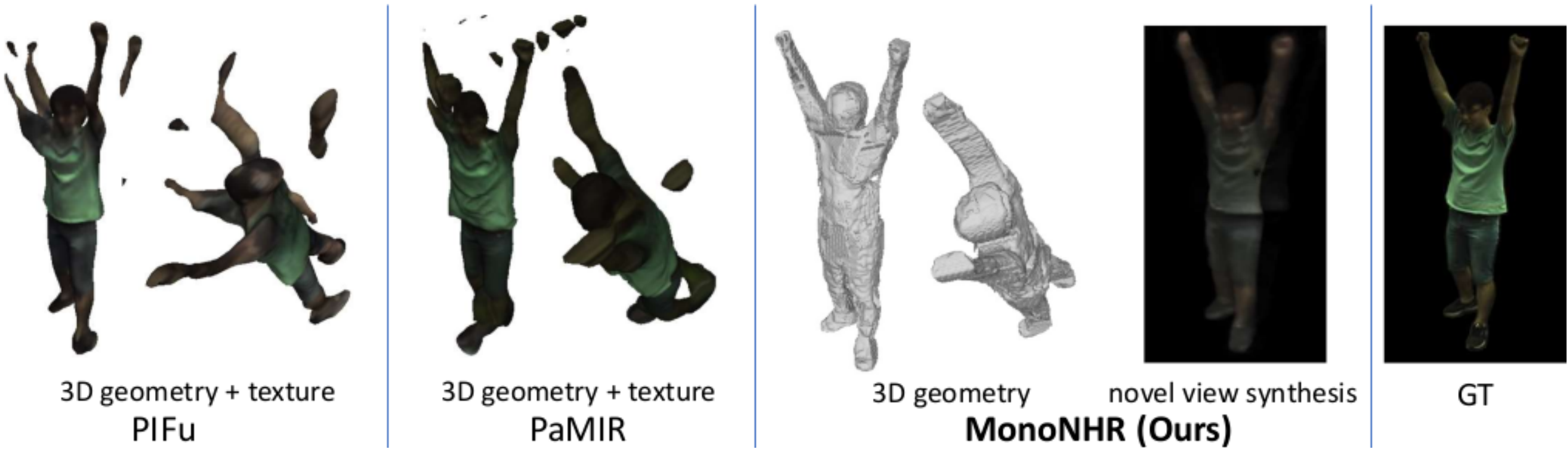}
\end{center}
\vspace*{-5mm}
\caption{
Comparison between state-of-the-art non-NeRF methods and ours on ZJU-MoCap.
}
\vspace*{-2mm}
\label{fig:sota_cape}
\end{figure}

\noindent\textbf{Comparison with non-NeRF methods.}
Figure~\ref{fig:sota_cape} shows that MonoNHR produces much better qualitative results than PIFu~\cite{saito2019pifu} and PaMIR~\cite{zheng2021pamir} on ZJU-Mocap.
Note that it is difficult to measure their PSNR/SSIM due to the pixel-level misalignment between their rendered images and the GT ones.
Table~\ref{table:sota_cape} shows that ours achieves much better CD on CAPE~\cite{ma2020learning}.
The results of PIFu and PaMIR are obtained by running their officially released codes and pre-trained weights.
We tested with MonoNHR trained on AIST.

\subsection{Ablation studies}

\begin{figure}[t]
\begin{center}
\begin{subfigure}{1.0\linewidth}
\includegraphics[width=1.0\linewidth]{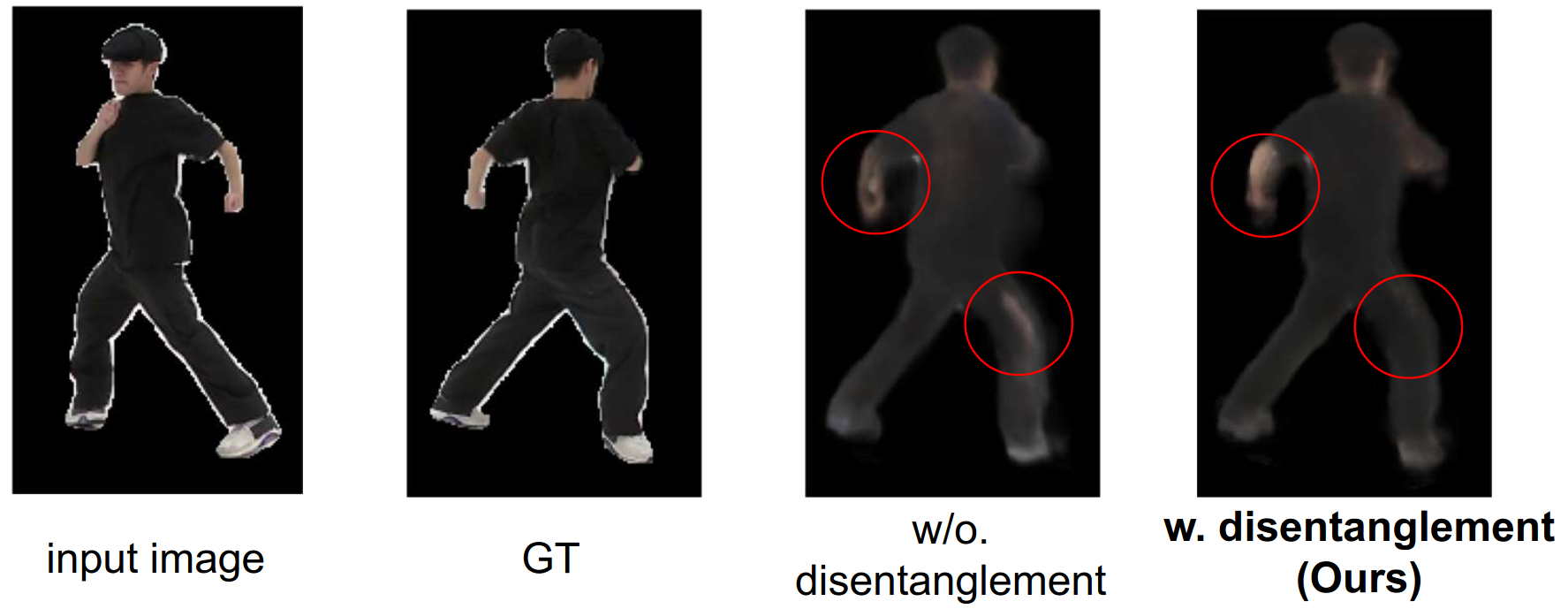}
\vspace*{-5mm}
\caption{}
\label{fig:ablation_disentangle}
\end{subfigure}
\begin{subfigure}{1.0\linewidth}
\includegraphics[width=\linewidth]{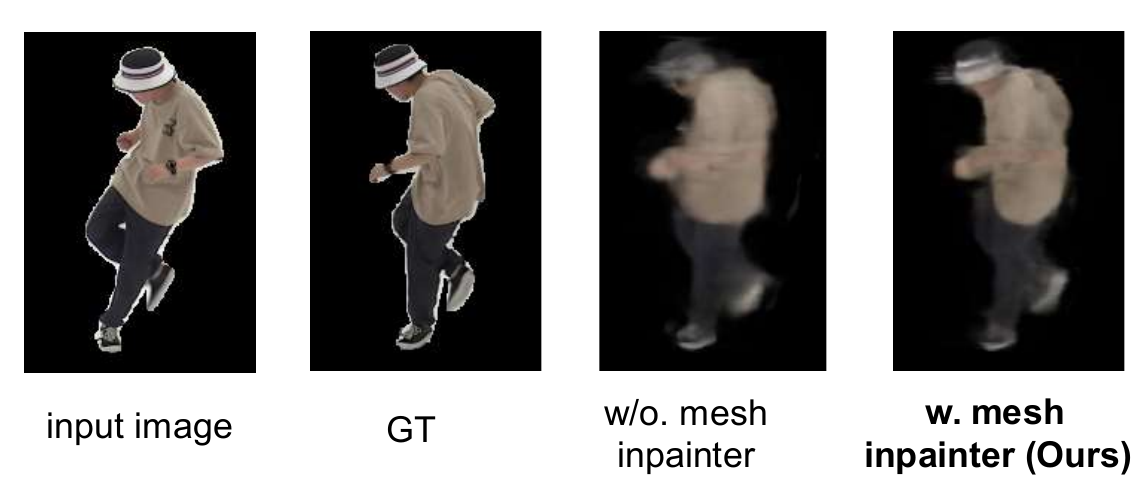}
\vspace*{-5mm}
\caption{}
\label{fig:ablation_mesh_inpainter}
\end{subfigure}
\begin{subfigure}{1.0\linewidth}
\includegraphics[width=\linewidth]{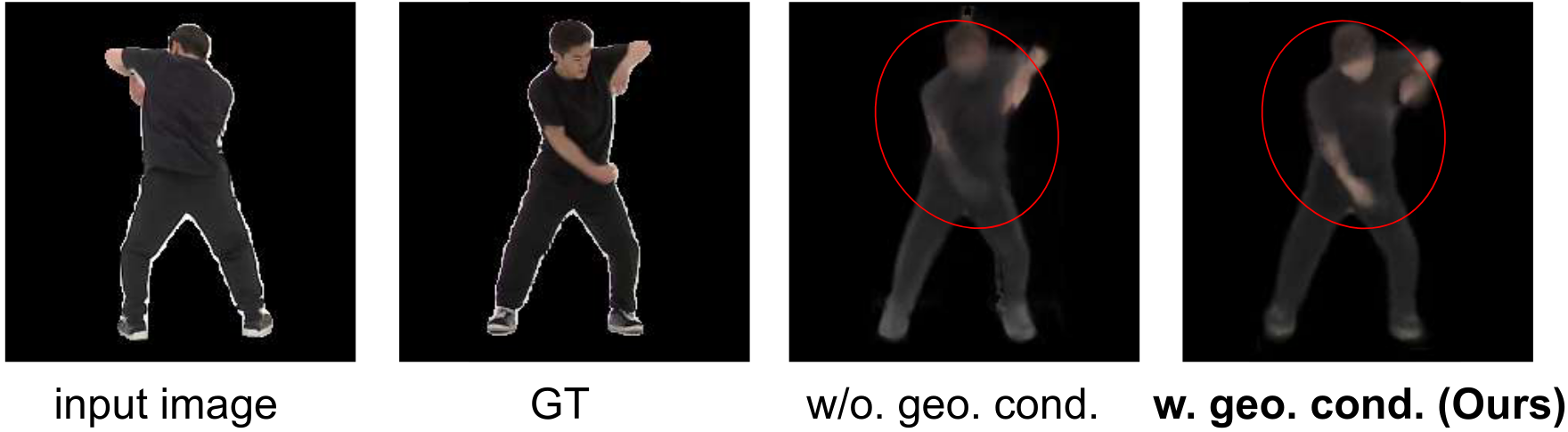}
\vspace*{-5mm}
\caption{}
\label{fig:ablation_geo_condition}
\end{subfigure}
\end{center}
\vspace*{-5mm}
\caption{
Qualitative results for all three ablation studies, on examples taken from ZJU-MoCap and AIST
}
\vspace*{-4mm}
\label{fig:ablation}
\end{figure}

\noindent\textbf{Disentanglement of geometry and texture.}
To show the benefit of the proposed disentanglement, we design a model where voxel features $\mathbf{v}$ are fed to both geometry and texture branches.
Results confirm that disentangling geometry and texture improves predictions both qualitatively and quantitatively (Table~\ref{table:ablation}).
As shown in Figure~\ref{fig:ablation_disentangle}, noise on the invisible surface's texture is removed, and the overall visual quality is enhanced.




\begin{table}[t]
\small
\centering
\setlength\tabcolsep{1.0pt}
\def\arraystretch{1.1}
\begin{tabular}{C{3.0cm}|C{1.1cm}}
\specialrule{.1em}{.05em}{.05em}
method & CD $\downarrow$ \\ \hline
PIFu~\cite{saito2019pifu} & 79.67 \\
PaMIR~\cite{zheng2021pamir} & 74.24 \\ 
\textbf{MonoNHR (Ours)} &  \textbf{23.94}\\
\specialrule{.1em}{.05em}{.05em}
\end{tabular}
\vspace*{-3mm}
\caption{
{Comparison between state-of-the-art non-NeRF methods and ours on CAPE.} 
}
\label{table:sota_cape}
\end{table}

\noindent\textbf{Mesh Inpainter.}
Using the Mesh Inpainter improves results both qualitatively (Fig.~\ref{fig:ablation_mesh_inpainter}) and quantitatively (Table~\ref{table:ablation}).
The critical challenge of the monocular setting is inferring the geometry and texture of invisible areas.
To this end, the backbone should produce proper features for unseen parts based on the visible ones, which amounts to learning human priors, such as symmetry of human parts and color similarity between surfaces belonging to the same part.
While these could be learned by the rendering loss at the end of the network, we tried to facilitate it by giving a more direct signal to the backbone using mesh inpainting.

\begin{table}[t]
\small
\centering
\setlength\tabcolsep{1.0pt}
\def\arraystretch{1.1}
\begin{tabular}{C{4.8cm}|C{1.1cm}|C{1.1cm}}
\specialrule{.1em}{.05em}{.05em}
setting & PSNR $\uparrow$ & SSIM $\uparrow$ \\ \hline
w/o Mesh Inpainter & 17.52 & 0.7064  \\ 
w/o disentanglement & 17.53 & 0.7167  \\
w/o geo. cond. & 17.38 &  0.6634 \\
\textbf{Ours: full model} &  \textbf{17.61} & \textbf{0.7186} \\ \specialrule{.1em}{.05em}{.05em}
\end{tabular}
\vspace*{-3mm}
\caption{
{Ablation studies on AIST.}
}
\label{table:ablation} 
\end{table}


\noindent\textbf{Geometry-conditioned texture estimation.}
Conditioning on geometry when estimating texture is essential (Fig.~\ref{fig:ablation_geo_condition} and Table~\ref{table:ablation}).
Indeed, geometry provides information on whether a 3D query point is occupied or not, and such occupancy information is important to produce accurate textures at correct locations.

\subsection{Qualitative results} \label{subsec:qualitative_results}
\begin{figure}[t]
\begin{center}
\includegraphics[width=0.22\linewidth]{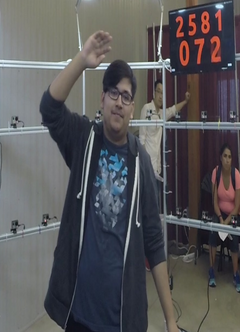}
\includegraphics[width=0.22\linewidth]{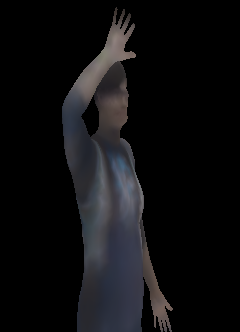}
\includegraphics[width=0.22\linewidth]{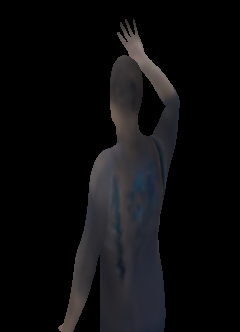} 
\includegraphics[width=0.22\linewidth]{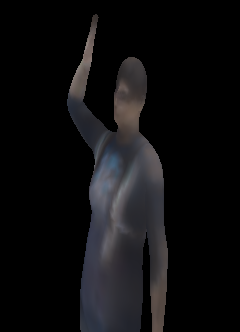}\\
\includegraphics[width=0.22\linewidth]{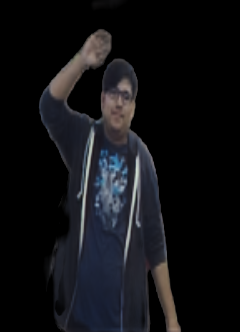}
\includegraphics[width=0.22\linewidth]{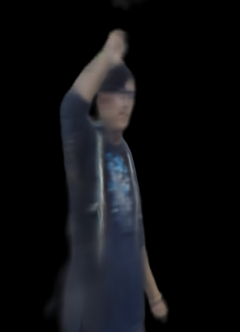}
\includegraphics[width=0.22\linewidth]{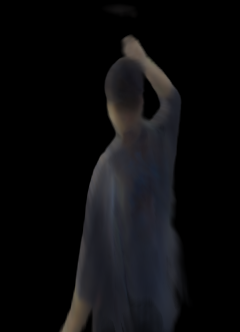} 
\includegraphics[width=0.22\linewidth]{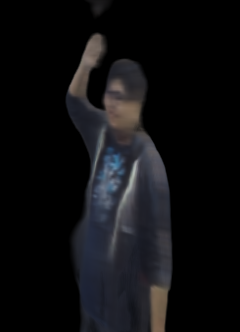}\\

\includegraphics[width=0.22\linewidth]{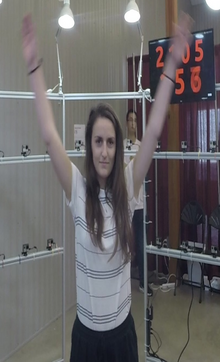}
\includegraphics[width=0.22\linewidth]{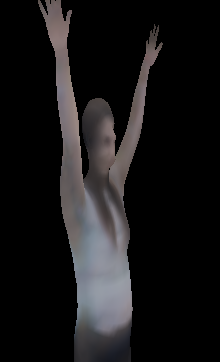}
\includegraphics[width=0.22\linewidth]{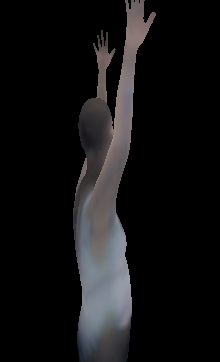} 
\includegraphics[width=0.22\linewidth]{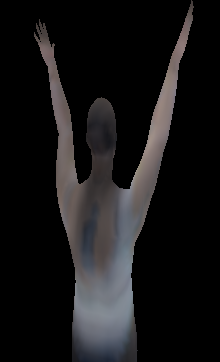}\\
\includegraphics[width=0.22\linewidth]{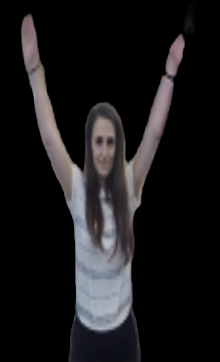}
\includegraphics[width=0.22\linewidth]{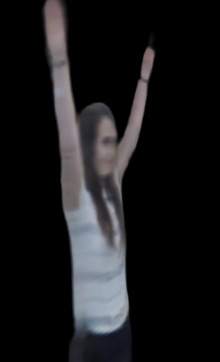}
\includegraphics[width=0.22\linewidth]{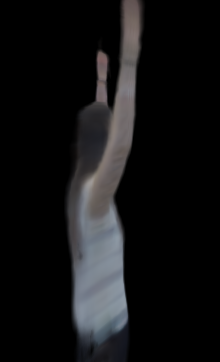} 
\includegraphics[width=0.22\linewidth]{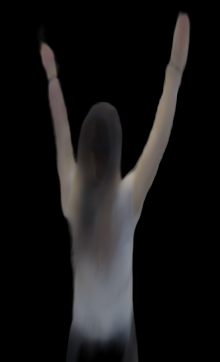}\\
\end{center}
\vspace*{-5mm}
\caption{
Inpainting of  SMPL mesh (top) vs MonoNHR (bottom)  for  input images  from HUMBI (top left).  
}
\vspace*{-4mm}
\label{fig:humbi_inpaint_comparison}
\end{figure}

The HUMBI dataset~\cite{yu2020humbi} has a high diversity of human subjects in terms of body shape, age, ethnicity, clothing and accessories. We train on this data in order to evaluate the robustness and generalization capabilities of our method.
Figures~\ref{fig:banner} and \ref{fig:humbi_inpaint_comparison} show some results on test subjects unseen during training. In the latter, we show the benefits of using neural rendering by comparing with surface renderings of the inpainted SMPL mesh. Despite the accuracy of the SMPL annotations on this dataset, we can clearly see the inherent limits of texture based approaches. They typically fail in regions where the true surface largely differs from the template shape which is often the case \eg with hair or wide cloths. In contrast, even though our approach makes use of the SMPL vertices, we exhibit good robustness to these cases.

\section{Conclusion}
We proposed MonoNHR, a NeRF-based approach that renders robust free-viewpoint images of an arbitrary human given a single monocular image. 
By disentangling 3D geometry features from texture features and enforcing the feature extractor to exploit human priors (\textit{e.g.}, symmetry), we reached state-of-the-art performance for the problem of novel view synthesis of an arbitrary person from a monocular observation. 
We also showed quantitative and qualitative results using estimated meshes, which take a step toward the authentic monocular setting.
We believe our results are inspiring in that 1) robust view synthesis for partially observed humans is feasible with a NeRF-based approach, and
2) we tackles the monocular problem without explicit 3D supervision, and thus it can be more easily scaled-up to large datasets than methods that rely on 3D scans.
In future work, adversarial losses could be employed to learn human priors further and improve the realism of novel view synthesis.

{\small
\bibliographystyle{ieee_fullname}
\bibliography{main}
}

\clearpage
\twocolumn[{
\begin{center}
\begin{Large}
\textbf{Supplementary Material \textit{for} \\ \vspace{2mm}
MonoNHR: Monocular Neural Human Renderer}

\end{Large}
\end{center}
\vspace*{+2em}
}]

In this supplementary material, we first analyze the additional qualitative results in Section~\ref{supp:quality} (\textbf{Please refer to the linked videos.}).
Then, we provide the implementation details of MonoNHR, pixelNeRF, and NHP that are adapted to a single image input for reproducibility in Section~\ref{supp:rep}.
In Section~\ref{supp:data}, we explain detailed training and testing protocols of ZJU-MoCap and AIST that we used in our experiments. 
\section{Qualitative results}
\label{supp:quality}
We provide a video\footnote{\url{https://youtu.be/9-hfGf7dRw4}} showing more qualitative results and comparisons on the ZJU-MoCap dataset~\cite{peng2021neural}, as well as the HUMBI dataset~\cite{yu2020humbi}. Below are more details and comments regarding the video.
\subsection{Results on ZJU-MoCap}
We show video results on ZJU-MoCap, both on dynamic scenes (where the input frame changes for each rendered image) and static scenes (where a video is generated given a single input image). Comparisons with NHP~\cite{kwon2021neural} show that our method is quite stable with respect to time, and does not suffer from the jittering effect present in the NHP results.
Other results show that MonoNHR can better render loose clothing, suggesting that it is able to correctly infer volumetric information around the underlying body-model, whereas NHP renderings tend to stick to the body-model.
Finally, in some cases, NHP has trouble generalizing to a new subject given only a single image. 

\subsection{Results on HUMBI} \label{supp:humbi}
In order to challenge the generalization capabilities of MonoNHR, we also train and test it on the HUMBI dataset~\cite{yu2020humbi}, which has a high diversity of human subjects in terms of body shape, age, race, clothing and accessories. We train our method on 9822 frames, taken from 334 subjects, with various poses. Qualitative results on diverse test subjects and poses are included in the attached video. They demonstrate that our method generalizes well to new people, unseen during training.
Also, some interesting effects can be noted in the attached video. 
For example, our method learns to correctly reconstruct long hair in the back, given only a single frontal view of a person.

\begin{figure*}[!t]
\begin{center}
\includegraphics[width=0.24\linewidth]{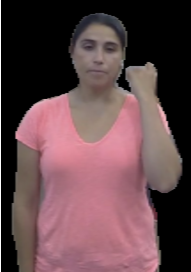}
\includegraphics[width=0.24\linewidth]{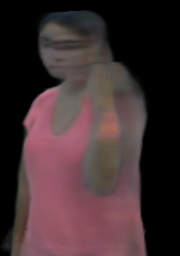}
\includegraphics[width=0.24\linewidth]{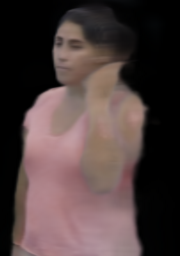}
\includegraphics[width=0.24\linewidth]{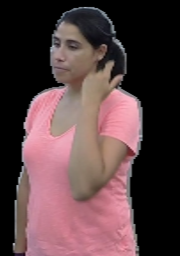} \\
\includegraphics[width=0.24\linewidth]{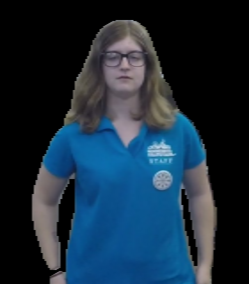}
\includegraphics[width=0.24\linewidth]{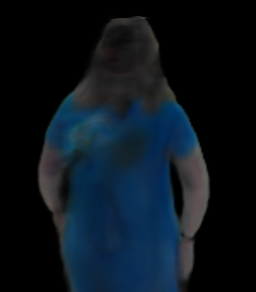}
\includegraphics[width=0.24\linewidth]{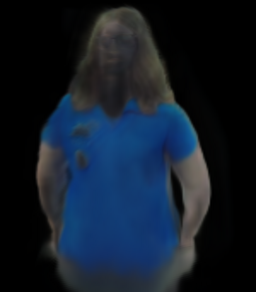}
\includegraphics[width=0.24\linewidth]{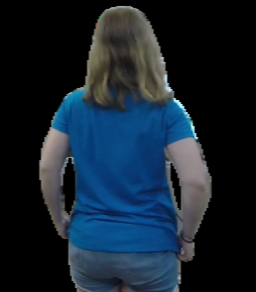} \\
\begin{subfigure}{0.24\linewidth}
\includegraphics[width=1.0\linewidth]{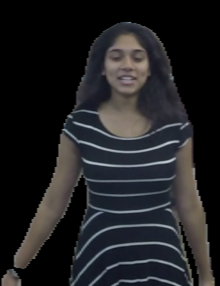}
\caption{Input image}
\end{subfigure}
\begin{subfigure}{0.24\linewidth}
\includegraphics[width=1.0\linewidth]{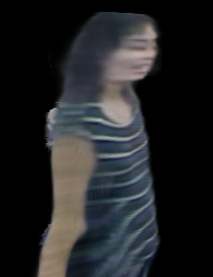}
\caption{NHP}
\end{subfigure}
\begin{subfigure}{0.24\linewidth}
\includegraphics[width=1.0\linewidth]{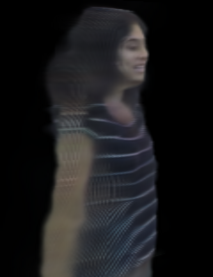}
\caption{Ours}
\end{subfigure}
\begin{subfigure}{0.24\linewidth}
\includegraphics[width=1.0\linewidth]{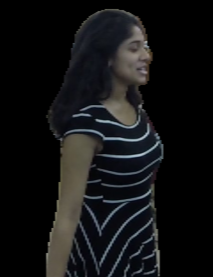}
\caption{Ground Truth}
\end{subfigure}
\end{center}
\caption{
Comparison with NHP on HUMBI~\cite{yu2020humbi}, qualitative results. From left to right: input image, NHP, ours, ground truth
}
\label{fig:nhp_humbi}
\end{figure*}

We retrain NHP on HUMBI for an additional comparison. We use the same protocol as MonoNHR (see section \ref{supp:humbi}). Implementation details are given in section \ref{supp:nhp}.
Table \ref{table:nhp_humbi} shows the average numerical results on 13 test subjects, unseen during training. (For each subject, we test on 5 views of one frame).
Figure \ref{fig:nhp_humbi} shows some qualitative comparisons on some of those test subjects.
Our method produces more realistic and visually pleasing renderings. However, like NHP, it still suffers from artifacts for rendering the opposite surface of an observed view, as can be seen in the second row. These come from the difficulty of extracting proper features for novel views only from observed image features.

\begin{table}[t]
\small
\centering
\setlength\tabcolsep{1.0pt}
\def\arraystretch{1.1}
\begin{tabular}{C{3.0cm}|C{1.1cm}|C{1.1cm}}
\specialrule{.1em}{.05em}{.05em}
method & PSNR $\uparrow$ & SSIM $\uparrow$ \\ \hline
NHP~\cite{kwon2021neural} & 17.472 & 0.7167\\ 
\textbf{MonoNHR (Ours)} &  \textbf{19.750} & \textbf{0.7463}\\
\specialrule{.1em}{.05em}{.05em}
\end{tabular}
\caption{
{Comparison with NHP~\cite{kwon2021neural} on HUMBI.} 
}
\label{table:nhp_humbi}
\end{table}

\section{Reproducibility}
\label{supp:rep}
\subsection{MonoNHR}

\noindent\textbf{Backbone.}
The backbone takes the input image $\mathbf{I} \in \mathbb{R}^{3 \times 256 \times 256}$ and produces the feature map $\mathbf{F} \in \mathbb{R}^{64 \times 128 \times 128}$.
We use ResNet18~\cite{he2016deep} from which we removed the classifier head, as MonoNHR's backbone network architecture.
Following pixelNeRF~\cite{yu2021pixelnerf}, the backbone extracts multiple feature maps at different scales, right after the first pooling operation of ResNet18 and after ResNet 3 layers.
The multi-scale feature maps have the following shapes (channel dimension $\times$ height $\times$ width) :
\\
1. $64 \times 128 \times 128$\\
2. $64 \times 64 \times 64$\\
3. $128 \times 32 \times 32$\\
4. $256 \times 16 \times 16$\\

We upsample the second, third, and fourth feature maps to the first feature map's height and width with bilinear interpolation, and concatenate all the feature maps along the channel dimension.
Then, we feed the concatenated feature map with 512 channel dimension to a 1-by-1 convolution layer to reduce the channel dimension, which outputs the feature map $\mathbf{F}$.

\begin{table*}[!t]
\centering
\scalebox{0.9}{
\centering
\begin{tabular}{c|c|c}
\hline
& layer description & output dimensions  \\ \hline \hline
layer index & input volume & D $\times$ H $\times$ W $\times$ (64+1) \\ \hline
1-2 & (kernel size 3, out channel 32, stride 1, dilation 1) $\times$ 2 & D $\times$ H $\times$ W $\times$ 32 \\
3 & (kernel size 3, out channel 32, stride 2, dilation 1) & D/2 $\times$ H/2 $\times$ W/2 $\times$ 32 \\
4-5 & (kernel size 3, out channel 32, stride 1, dilation 1) $\times$ 2 & D/2 $\times$ H/2 $\times$ W/2 $\times$ 32 \\
6 & (kernel size 3, out channel 32, stride 2, dilation 1) & D/4 $\times$ H/4 $\times$ W/4 $\times$ 32 \\
7-9 & (kernel size 3, out channel 32, stride 1, dilation 2) $\times$ 3 & D/4 $\times$ H/4 $\times$ W/4 $\times$ 32 \\
10 & (kernel size 3, out channel 64, stride 2, dilation 1) & D/8 $\times$ H/8 $\times$ W/8 $\times$ 64 \\
11-13 & (kernel size 3, out channel 64, stride 1, dilation 2) $\times$ 3 & D/8 $\times$ H/8 $\times$ W/8 $\times$ 64 \\
14 & (kernel size 3, out channel 64, stride 2, dilation 1) & D/16 $\times$ H/16 $\times$ W/16 $\times$ 64 \\
15-17 & (kernel size 3, out channel 64, stride 1, dilation 2) $\times$ 3 & D/16 $\times$ H/16 $\times$ W/16 $\times$ 64 \\
\hline
\end{tabular}
} 
\vspace{3mm}
\caption{Architecture of the geometry branch's sparse 3D CNN. Each convolution layer is followed by 1D batch normalization and ReLU.}
\label{tab:sparse_3d_cnn}
\end{table*}

\begin{figure}
\begin{center}
\includegraphics[width=0.5\linewidth]{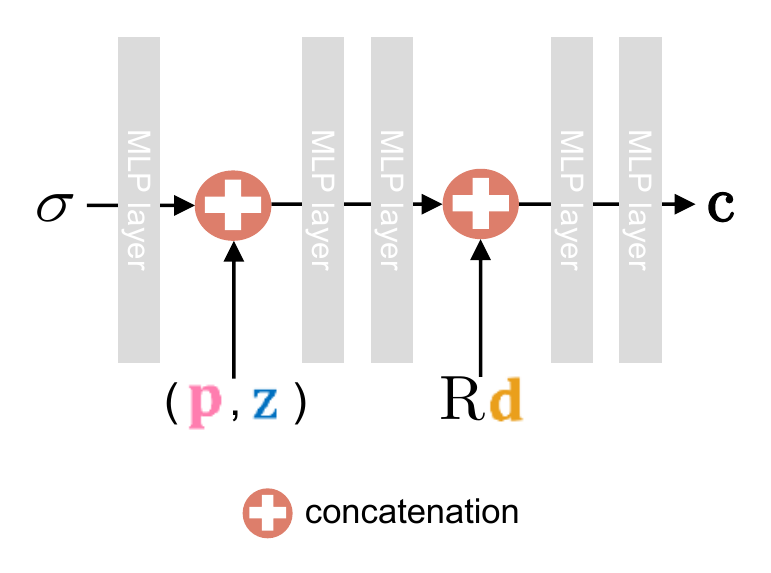}
\end{center}
\caption{
Architecture of the texture branch. $\mathbf{R}$ denotes the world-to-camera rotation matrix.
}
\label{fig:texture_branch}
\end{figure}

\noindent\textbf{Mesh inpainter.}
The mesh inpainter takes the mesh feature $\mathbf{M} \in \mathbb{R}^{M \times (64+1)}$ as input and estimates the vertices' colors $\hat{\mathbf{K}} \in \mathbb{R}^{M \times 3}$, where $M$ is the number of SMPL~\cite{loper2015smpl} mesh vertices, which is 6890.
The mesh inpainter consists of two MLP layers with a ReLU~\cite{nair2010rectified} activation between them.
The first layer reduces the channel dimension from (64+1) to 32, and the second layer reduces it to 3, which corresponds to RGB channels.

\noindent\textbf{Geometry branch - Sparse 3D convolution.} 
The geometry branch first builds a sparse 3D volume with the mesh feature $\mathbf{M}$, given a SMPL mesh similarly to NB~\cite{peng2021neural} and NHP~\cite{kwon2021neural}. 
We obtain the 3D bounding box of the SMPL mesh in the input camera's coordinate system with $5cm$ padding along the depth, height, and width dimensions.
The 3D bounding box is transformed into a discretized volume with a voxel size of $5mm \times 5mm \times 5mm$, where the depth, height, and width ($D \times H \times W$) depend on the SMPL mesh.
We calculate the discretized coordinates of mesh vertices, which correspond to the discretized volume's voxels, and fill the voxels with $\mathbf{M}$ to get the sparse 3D volume.
The sparse 3D volume is fed to a sparse 3D CNN~\cite{graham2018spconv}. 
We describe the architecture of this sparse 3D CNN in Table~\ref{tab:sparse_3d_cnn}.
The output features of the 5th, 9th, 13th, and 17th layers are the mesh volumes defined in Section 3.3 of the main manuscript.
We sample the features of 3D query points from each mesh volume and concatenate the sampled features to get the voxel feature $\mathbf{v} \in \mathbb{R}^{192}$, which is the geometry-dedicated feature.

\noindent\textbf{Geometry branch - MLP.} 
The voxel feature $\mathbf{v}$, the pixel feature $\mathbf{p}$, and the root joint-relative depth $\mathrm{z}$ (see the main manuscript's Section 3.1 for the definition of  $\mathbf{p}$ and $\mathrm{z}$) are concatenated and fed to the MLP network, that consists of four layers, where each layer is followed by ReLU activation except the last one.
The first 3 layers produce features with a channel dimension 256, and the last layer regresses the density $\sigma$ of the 3D query point.

\begin{algorithm}
\caption{Pseudocode of MonoNHR in a PyTorch-style}
\label{alg:code}
\definecolor{codeblue}{rgb}{0.25,0.5,0.5}
\lstset{
  backgroundcolor=\color{white},
  basicstyle=\fontsize{7.2pt}{7.2pt}\ttfamily\selectfont,
  columns=fullflexible,
  breaklines=true,
  captionpos=b,
  commentstyle=\fontsize{7.2pt}{7.2pt}\color{codeblue},
  keywordstyle=\fontsize{7.2pt}{7.2pt},
}
\begin{lstlisting}[language=python]
# data: mini-batch 
# x_c: 3D query point x's camera coordinates. Nx(num_raysx64)x3.
# mesh_c: mesh vertices' camera coordinates. Nx6890x3.
# root_j_c: root joint's camera coordinates. Nx1x3.
# K: intrinsic camera matrix. Nx3x3.
# viewdir_c: 3D query point's rendering view direction in the input camera's coordinate system. Nx(num_raysx64)x3. (d).
# N: mini-batch size

""" Data loader """
img, x_c, mesh_c, root_j_c, K, viewdir_c = 
data['img'], data['x_c'], data['mesh_c'], 
data['root_j_c'], data['K'], data['viewdir_c']


""" Backbone """
# featmap: Nx64x128x128. (F).
featmap = backbone.forward(img)

# mesh_feat: Nx6890x(64+1). (M).
mesh_i = projection(mesh_c, K)
mesh_i_feat = bilinear_sample(featmap, mesh_i)
mesh_z = mesh_c[:, :, 2:] - root_j_c[:, :, 2:]
mesh_feat = concat([mesh_i_feat, mesh_z], dim=2)

# pixel_feat: Nx(num_raysx64)x64. (p).
# x_z: Nx(num_raysx64)x1. (z).
x_i = projection(x_c, K)
pixel_feat = bilinear_sample(featmap, x_i)
x_z = x_c[:, :, 2:] - root_j_c[:, :, 2:]


""" Mesh inpainter """
# mesh_rgb_pred: Nx6890x3. (\hat{K}).
mesh_rgb_pred = mesh_inpainter.forward(mesh_feat)


""" Geometry branch """
# in_volume: Nx(64+1)xDxHxW. (sparse).
in_volume = encode_volume(mesh_feat, mesh_c)
mesh_volumes = sparse_3dconv.forward(in_volume)

# voxel_feat: Nx(num_raysx64)x192. (v).
voxel_feat =  trilinear_sample_and_concat(mesh_volumes, x_c)

# sigma: Nx(num_raysx64)x1. (\sigma(x)).
sigma = geo_mlp.forward(voxel_feat, pixel_feat, x_z) # regress density of x


""" Texture branch """
# rgb: Nx(num_raysx64)x3. (c(x)).
rgb = tex_net.forward(pixel_feat, x_z, viewdir_c, sigma) # regress RGB of x, conditioning on sigma

\end{lstlisting}
\end{algorithm}

\noindent\textbf{Texture branch.}
The texture branch takes the pixel feature $\mathbf{p}$, the root joint-relative depth $\mathrm{z}$, and the target view's ray direction $\mathbf{d}$, and the predicted density value $\sigma$ as input, and regresses the RGB value $\mathbf{c}$ of the 3D query point. 
It is also an MLP network of five layers, each layer being followed by ReLU activation except the last one.
Figure~\ref{fig:texture_branch} presents the architecture of the texture branch.

\noindent\textbf{Pseudocode.}
Algorithm~\ref{alg:code} provides the pseudocode of MonoNHR's forward pass during training.
As mentioned in the main manuscript's Section 4, a batch of 1024 rays is sampled from each rendering target image.
Then, 64 points are sampled from each ray as explained in Section 3.5.
As a result, for each iteration of training, the total number of (number of rays by 64) 3D query points' density and RGB values are regressed. 

\subsection{pixelNeRF}
We trained and tested pixelNeRF~\cite{yu2021pixelnerf} on ZJU-MoCap~\cite{peng2021neural} and AIST~\cite{li2021learn,tsuchida2019aist} using the official implementation.
We used the model setting for DTU MVS~\cite{jensen2014large} benchmark.
It uses ResNet34~\cite{he2016deep} as a backbone, which is heavier than that of MonoNHR.
To achieve a fair comparison to MonoNHR, we used the same volume rendering technique described in the main manuscript's Section 3.5.
For example, the 3D query points are bounded by the given SMPL mesh's 3D bounding box.
Also, we normalized the translation of the 3D query points' coordinates, which are the input to the pixelNeRF network, by aligning them with the root joint of the given SMPL mesh.
This translation normalization significantly improved the results of pixelNeRF.

\subsection{NHP} \label{supp:nhp}
We adapted NHP~\cite{kwon2021neural} to take a monocular image as input.
We re-implemented NHP from the information available in the paper, the implementation of NB~\cite{peng2021neural} and pixelNeRF~\cite{yu2021pixelnerf}, following the instructions of the authors of NHP~\cite{kwon2021neural}.

Again, in order to achieve a fair comparison with MonoNHR, we used the same volume rendering technique described in the main manuscript's Section 3.5.

\section{Detailed training and testing protocols}
\label{supp:data}
\noindent\textbf{ZJU-MoCap.}
We use subjects number 313, 377, 392 as testing set, and the remaining 7 subjects as training set. 
We uniformly sample 60 and 10 frames from each subject's video, during training and testing respectively.

\noindent\textbf{AIST.}
We follow the official split protocol for training and testing, and convert videos to images under exact 60 fps (frames per second) referring to the official homepage\footnote{\url{https://google.github.io/aistplusplus\_dataset/download.html}}.
We uniformly sub-sample a tenth of the extracted frames both for training and testing.
Since AIST does not provide human masks, we obtain them using the third party implementation\footnote{\url{https://github.com/Engineering-Course/CIHP\_PGN}} of PGN~\cite{gong2018instance}.
In our experiments, we used the model trained on the CIHP dataset~\cite{gong2018instance}.

\section{Inputs of the density estimation MLP}

\begin{figure*}[!t]
\begin{center}
\includegraphics[width=0.9\linewidth]{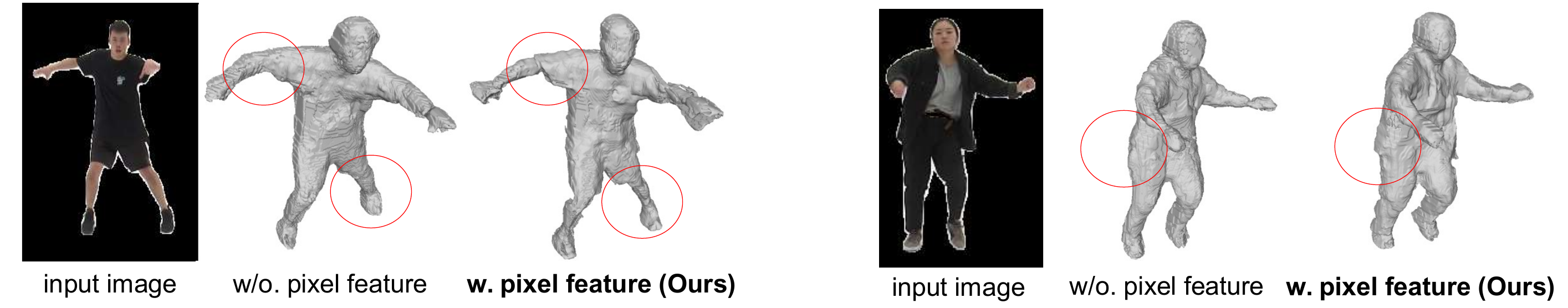}
\end{center}
\vspace*{-7mm}
\caption{
Comparison between models that take different inputs in their density estimation MLP.
}

\vspace*{-5mm}
\label{fig:ablation_geo_input}
\end{figure*}

Figure~\ref{fig:ablation_geo_input} shows that our inputs to the density estimation MLP, namely voxel feature $\mathbf{v}$, pixel feature $\mathbf{p}$, and root joint-relative depth $\mathrm{z}$, clearly reconstruct better 3D shapes.
Providing only a voxel feature $\mathbf{v}$ to the density estimation MLP, the network fails to distinguish the 3D geometry of clothes from the body. 
Since the mesh vertex positions provide 3D human naked body shape information
, it lacks clothed human shape information.
Therefore, by solely using a voxel feature for the density estimation, MonoNHR can struggle to recover accurate clothed human shapes, especially when the input human is wearing loose clothes. 
Additional pixel features with root joint-relative depth can provide such cloth silhouette information.

\section{Interest of mesh inpainter}

\begin{figure*}[!t]
\begin{center}
\includegraphics[width=0.9\linewidth]{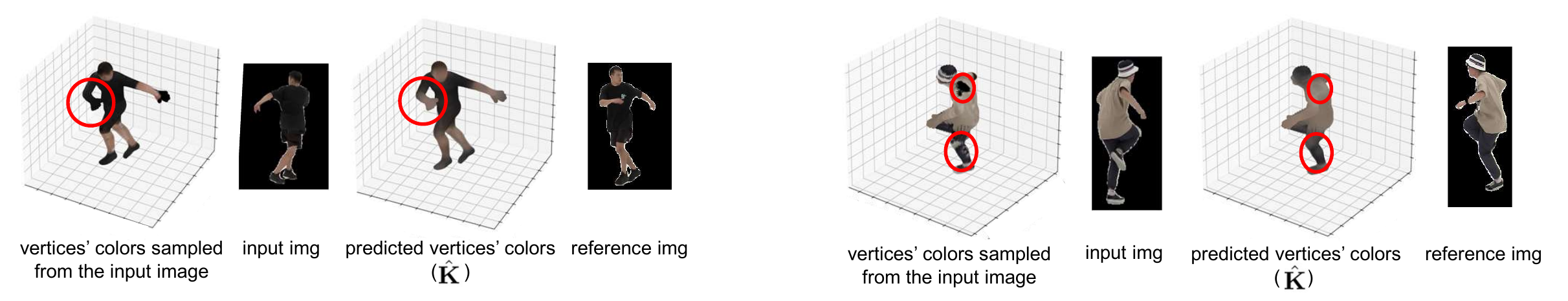}
\end{center}
\caption{
Inpainted SMPL meshes, w/ and w/o mesh inpainter.
}

\label{fig:mesh_inpainter}
\end{figure*}

Figure~\ref{fig:mesh_inpainter} shows the output of the mesh inpainter, compared to a naive back-projection of image colors onto the SMPL mesh. Results show the mesh inpainter is able to predict realistic colors for occluded limbs, and thus, indirectly contributes to conditioning good mesh features for the subsequent 3d rendering task, as our ablation study shows in the main paper.
\section{Implementation detail}
We implement our approach using PyTorch~\cite{paszke2017automatic} and the pseudo-codes are provided in the supplementary material.
During training, we randomly sample 3-4 novel view images and use them as the rendering targets.
A batch of 1024 rays is sampled from each target and fed to MonoNHR.
The weights of MonoNHR are updated using the Adam optimizer~\cite{kingma2014adam}.
The initial learning rate is set to $5\times10^{-4}$ and decays exponentially to $5\times10^{-5}$ following Neural Body~\cite{peng2021neural}.
We use one 2080 Ti GPU for training and testing.
\vspace{-1mm}
\section{Limitations}
\label{supp:limitations}
\vspace{-1mm}
Although the proposed MonoNHR highly outperforms the recent generalizable NeRFs~\cite{yu2021pixelnerf,kwon2021neural} adapted for monocular single input images, there are limitations to be resolved in future work.
First, though it is inevitable, the rendered image quality of invisible surfaces is not as realistic as the visible surfaces.
While the overall geometry and colors are correct, artifacts remain, such as blur and subtle color intensity differences.
Second, 3D geometry reconstruction depends on the given SMPL mesh. 
MonoNHR recovers relatively robust and detailed 3D shape of the human compared with NHP~\cite{kwon2021neural}, but the 3D shape accuracy still degrades if the given SMPL mesh's 3D pose is inaccurate.
Third, occluded human parts' textures are still occasionally rendered with occluding human parts' textures.

To resolve the first and third limitations, adversarial supervision could be applied.
For instance, a discriminator comparing the input image and the rendered image could be designed to enforce consistency.
Also, one could explicitly model the lighting condition during rendering to improve realism.
Regarding the second limitation, one could attempt to jointly refine the SMPL parameters with the rendering loss, as suggested by several previous methods~\cite{kwon2021neural,peng2021animatable,wu2020multi}.

\end{document}